\documentclass[10pt,twocolumn,letterpaper]{article}

\usepackage{cvpr}              

\usepackage{graphicx}
\usepackage{amsmath}
\usepackage{amssymb}
\usepackage{booktabs}
\usepackage{url}            
\usepackage{booktabs}       
\usepackage{amsfonts}       
\usepackage{nicefrac}       
\usepackage{microtype}      
\usepackage{wrapfig}

\usepackage{amssymb}

\usepackage{amsmath}
\usepackage{amsfonts}
\usepackage{algorithm}
\usepackage{algorithmic}

\usepackage{times}
\usepackage{epsfig}
\usepackage{graphicx}
\usepackage{color}

\usepackage{multirow}
\usepackage{bm}

\usepackage{diagbox}

\usepackage{soul}
\usepackage{url}

\usepackage{float}

\usepackage{wrapfig,lipsum,booktabs}

\usepackage[pagebackref,breaklinks,colorlinks]{hyperref}

\usepackage[capitalize]{cleveref}
\crefname{section}{Sec.}{Secs.}
\Crefname{section}{Section}{Sections}
\Crefname{table}{Table}{Tables}
\crefname{table}{Tab.}{Tabs.}

\def\cvprPaperID{5389} 
\def\confName{CVPR}
\def\confYear{2022}

\begin{document}

\title{BaLeNAS: Differentiable Architecture Search via the Bayesian Learning Rule}

\author{Miao Zhang$^{1}$ Jilin Hu$^{1}$ Steven Su$^{2}$ Shirui Pan$^{3}$ Xiaojun Chang$^{4}$ Bin Yang$^{1}$  Gholamreza Haffari$^{3}$\\
$^{1}$Aalborg University \quad $^{2}$UTS\quad $^{3}$Monash University\quad $^{4}$RMIT University\\

}
\maketitle

\begin{abstract}
Differentiable Architecture Search (DARTS) has received massive attention in recent years, mainly because it significantly reduces the computational cost through weight sharing and continuous relaxation. However, more recent works find that existing differentiable NAS techniques struggle to outperform naive baselines, yielding deteriorative architectures as the search proceeds. Rather than directly optimizing the architecture parameters, this paper formulates the neural architecture search as a distribution learning problem through relaxing the architecture weights into Gaussian distributions. By leveraging the natural-gradient variational inference (NGVI), the architecture distribution can be easily optimized based on existing codebases without incurring more memory and computational consumption. We demonstrate how the differentiable NAS benefits from Bayesian principles, enhancing exploration and improving stability. The experimental results on NAS-Bench-201 and NAS-Bench-1shot1 benchmark datasets confirm the significant improvements the proposed framework can make. In addition, instead of simply applying the argmax on the learned parameters, we further leverage the recently-proposed training-free proxies in NAS to select the optimal architecture from a group architectures drawn from the optimized distribution, where we achieve state-of-the-art results on the NAS-Bench-201 and NAS-Bench-1shot1 benchmarks. Our best architecture in the DARTS search space also obtains competitive test errors with 2.37\%, 15.72\%, and 24.2\% on CIFAR-10, CIFAR-100, and ImageNet datasets, respectively.
\end{abstract}


\section{Introduction}
\label{sec1}

Neural Architecture Search (NAS) \cite{ren2020comprehensive,LiPYWLLC20,ChengZHDCDLG20,LiWWLLC21,LiTWPWLC21} is attaining increasing attention in the deep learning community by automating the labor-intensive and time-consuming neural network design process. More recently, NAS has achieved the state-of-the-art results on various deep learning applications, including image classification \cite{tan2019efficientnet}, object detection \cite{chen2019detnas}, stereo matching \cite{cheng2020hierarchical}. Although NAS has the potential to find high-performing architectures without human intervention, the early NAS methods have extremely-high computational requirements \cite{zoph2018learning,real2018regularized,guo2018irlas}. For example, in \cite{zoph2018learning,real2018regularized}, NAS costs thousands of GPU days to obtain a promising architecture through reinforcement
learning (RL) or evolutionary algorithm (EA). This high computational requirement in NAS is unaffordable for most researchers and practitioners. Since then, more researchers shift to improve the efficiency of NAS methods \cite{li2019random,guo2019single,pham2018efficient}. Weight sharing NAS, also called One-Shot NAS \cite{bender2018understanding,pham2018efficient}, defines the search space as a supernet, and only the supernet is trained for once during the architecture search. The architecture evaluation is based on inheriting weights from the supernet without retraining, thus significantly reducing the computational cost. \textit{Differentiable architecture search} (DARTS)  \cite{liu2018darts}, which is one of the most representative works, further relaxes the discrete search space into continuous space and jointly optimize supernet weights and architecture parameters with gradient descent, to further improve efficiency. Through employing two techniques, weight sharing \cite{bender2018understanding,pham2018efficient} and continuous relaxation \cite{liu2018darts,xie2018snas,GDAS,cai2018proxylessnas}, DARTS reformulates the discrete operation selection problem in NAS as a continuous magnitude optimization problem, which reduces the computational cost significantly and completes the architecture search process within several hours on a single GPU.

Despite notable benefits on computational efficiency from differentiable NAS, more recent works find it is still unreliable \cite{zela2019understanding,chen2020stabilizing} to directly optimize the architecture magnitudes. For example, DARTS is unable to stably obtain excellent solutions and yields deteriorative architectures during the search proceeds, performing even worse than random search in some cases \cite{sciuto2019evaluating}. This critical weakness is termed as \textit{instability} in differentiable NAS \cite{zela2019understanding}. Zela \etal~\cite{zela2019understanding} empirically point out that the instability of DARTS is highly correlated with the dominant eigenvalue of the Hessian of the validation loss with respect to the architectural parameters, while this dominant eigenvalue increases during the architecture search. Accordingly, they proposed a simple early-stopping criterion based on this dominant eigenvalue to robustify DARTS. In addition, Wang \etal~\cite{Rethinking2021} observe that the instability in DARTS's final discretization process of architecture selection, where the optimized magnitude could hardly indicate the importance of operations. On the other hand, several works \cite{shu2019understanding,chen2020drnas,li2019improving,zhang2020one} state that directly optimizing the architecture parameters without exploration easily entails the rich-gets-richer problem, leading to those architectures that converge faster at the beginning while achieve poor performance at the end of training, e.g. architectures with intensive \textit{skip-connections} \cite{chu2019fairnas,liang2019darts+}.

Unlike most existing works that directly optimize the architecture parameters, we investigate differentiable NAS from a distribution learning perspective, and introduce the \textbf{Ba}yesian \textbf{Le}arning rule \cite{khan2018fast,osawa2019practical,meng2020training,khan2020} to the architecture optimization in differentiable \textbf{NAS} with considering natural-gradient variational inference (NGVI) methods to optimize the architecture distribution, which we call \textbf{BaLeNAS}. We theoretically demonstrate how the framework naturally enhance the exploration for differentiable NAS and improves the stability, and the experimental results confirm that our framework enhances the performance for differentiable NAS. Rather than simply applying \textit{argmax} on the mean to get a discrete architecture, we for the first time leverage the training free proxies \cite{mellor2020neural,chen2021neural,abdelfattah2021zero} to select a more competitive architecture from the optimized distribution, without incurring any additional training costs. Specifically, our approach achieves state-of-the-art performance on NAS-Bench-201 \cite{BENCH102} and improves the performance on NAS-Bench-1shot1 \cite{zela2020nasbench1shot1} by large margins, and  obtains competitive results on CIFAR-10, CIFAR-100, and ImageNet datasets in the DARTS \cite{liu2018darts} search space, with test error 2.37\%, 15.72\%, and 24.2\%, respectively. Our contributions are summarized as follows.
\begin{itemize}
\item Firstly, this paper formulates the neural architecture search as a distribution learning problem and builds a generalized Bayesian framework for architecture optimization in differentiable NAS. We demonstrate that the proposed Bayesian framework is a practical solution to enhance exploration for differentiable NAS and improve  stability as a by-product via implicitly regularizing the Hessian norm. 

\item Secondly, instead of directly applying the \textit{argmax} on the learned parameters to get architectures, we for the first time leverage zero-cost proxies to select competitive architectures from the optimized distributions. As these proxies are calculated without any training, the architecture selection phase can be finished extremely efficiently.


\item Thirdly, the proposed framework is built based on DARTS and is also comfortable to be extended to other differentiable NAS methods with minimal modifications through leveraging the natural-gradient variational inference (NGVI). Experiments show that our framework consistently improves the baselines with obtaining more competitive architectures in various search spaces.

\end{itemize}

\section{Preliminaries}
\label{sec2}
\subsection{Differentiable Architecture Search}
\label{sec2.1}

Differentiable architecture search (DARTS) is built on weight-sharing NAS \cite{bender2018understanding,pham2018efficient}, where the supernet is trained for once per the architecture search cycle. Rather than using the heuristic methods \cite{pham2018efficient,zhang2020one} to search for the promising architecture in the discrete architecture space $\mathcal{A}$, DARTS \cite{liu2018darts} proposes the differentiable NAS framework by applying a continuous relaxation (usually a \textit{softmax}) to the discrete architecture space and enabling gradient descent for architecture optimization. Therefore, architecture parameters $\alpha_\theta$ and supernet weights $w$ could be jointly optimized during the supernet training, and the promising architecture parameters $\alpha_\theta^*$ are searched from the continuous search space $\mathcal{A}_\theta$ once the supernet is trained. The bilevel optimization formulation is usually adopted to alternatively learn $\alpha_\theta$ and $w$:
\begin{equation} \label{eq:bilevel}
\underset{\alpha_\theta \in \mathcal{A}_\theta}{\mathtt{min}} \ \mathcal{L}_{\mathtt{val}}\Big(\underset{w}{\mathtt{argmin}}\ \mathcal{L}_{\mathtt{train}}(w(\alpha_\theta),\alpha_\theta)\Big),
\end{equation}
and the best discrete architecture $\alpha^*$ is obtained after applying \textit{argmax} on $\alpha_\theta^*$. 

Despite notable benefits on computational efficiency from DARTS, more recent works find it is still unreliable \cite{zela2019understanding,chen2020stabilizing} that directly optimizes the architecture magnitudes, where DARTS usually observes a performance collapses with search progresses. This phenomenon is also called the instability of differentiable NAS  \cite{chen2020stabilizing}. Zela \etal~\cite{zela2019understanding} observed that the there is a strong correlation between the dominant eigenvalue of the Hessian of the validation loss and the architecture’s generalization error in DARTS, and keeping the the Hessian matrix's norm in a low level plays a key role in robustifying the performance of differentiable NAS \cite{chen2020stabilizing}. In addition, as described above, the differentiable NAS first relaxes the discrete architectures into continuous representations to enable the gradient descent optimization, and projects the continuous architecture representation $\alpha_{\theta}$ into discrete architecture $\alpha$ after the differentiable architecture optimization. However, more recent works \cite{Rethinking2021} cast doubts on the robustness of this discretization process in DARTS that the magnitude of architecture parameter $\alpha_{\theta}^*$ could hardly indicate the importance of operations with \textit{argmax}. Taking the DARTS as example, the searched architecture parameters $\alpha_{\theta}$ are continuous, while $\alpha$ is represented with \{0, 1\} after \textit{argmax}. DARTS assumes that the $\mathcal{L}_{val}(w^*,\alpha_{\theta}^*)$ is a good indicator to the validation performance of $\alpha$, $\mathcal{L}_{val}(w^*,\alpha^*)$. However, when we conduct the Taylor expansion on the local optimal $\alpha_{\theta}^*$ \cite{chen2020stabilizing,chen2020drnas}, we have:
\begin{equation} \label{eq:instability}
\resizebox{.9\linewidth}{!}{$
    \displaystyle
\begin{aligned}
\mathcal{L}_{val}(w^*,\alpha^*)&=\mathcal{L}_{val}(w^*,\alpha_{\theta}^*)+\triangledown_{\alpha_\theta}\mathcal{L}_{val}(w^*,\alpha_{\theta}^*)^T(\alpha^*-\alpha_{\theta}^*)\\
&+\frac{1}{2} (\alpha^*-\alpha_{\theta}^*)^{T}\mathcal{H}(\alpha^*-\alpha_{\theta}^*)\\
&=\mathcal{L}_{val}(w^*,\alpha_{\theta}^*)+\frac{1}{2} (\alpha^*-\alpha_{\theta}^*)^{T}\mathcal{H}(\alpha^*-\alpha_{\theta}^*)
\end{aligned}
$}
\end{equation}
where $\triangledown_{\alpha_\theta}\mathcal{L}_{val}=0$ due to the local optimality condition, and $\mathcal{H}$ is the Hessian matrix of $\mathcal{L}_{val}(w^*,\alpha_{\theta})$. We can see that the incongruence of the final continuous architecture representation and the final discrete architecture relates to the Hessian matrix's norm. However, as demonstrated by the empirical results in \cite{zela2019understanding}, the eigenvalue of this Hessian matrix increases during the architecture search, incurring more incongruence. 

\subsection{Bayesian Deep Learning}
\label{sec2.2}

Given a dataset $\mathcal{D}=\{\mathcal{D}_1, \mathcal{D}_1,...,\mathcal{D}_N\}$ and a deep neural network with parameters $\theta$, the most popular method to learn $\theta$ with $\mathcal{D}$ is Empricial Risk Minimization (ERM):
\begin{equation} \label{eq:erm}
\textup{min}\ \bar{\ell}(\theta):=\sum_{i=1}^{N}\ell_i(\theta)+\eta \mathcal{R}(\theta),
\end{equation}
where $\ell_i$ is a loss function, e.g., $\ell_i=-\textup{log}\ p(\mathcal{D}_i \mid \theta)$ for classification and $\mathcal{R}$ is the regularization term.

In contrast, the \textbf{Bayesian deep learning} estimate the posterior distribution of $\theta$, $p(\theta \mid \mathcal{D}):= p(\mathcal{D}\mid \theta )p(\theta )/p(\mathcal{D})$, where $p(\theta)$ is the prior distribution. However, the normalization constant $p(\mathcal{D})=\int p(\mathcal{D}\mid \theta )p(\theta )d\theta$ is difficult to compute for large DNNs. The variational inference (VI)  \cite{graves2011practical} resolves this issue in Bayesian deep learning by approximating $p(\theta \mid \mathcal{D})$ with a new distribution $q(\theta)$, and minimizes the Kullback-Leibler (KL) divergence between $p(\theta \mid \mathcal{D})$ and $q(\theta)$,
\begin{equation} \label{eq:minKL}
\textup{argmin}_\theta \textup{KL}(q(\theta)\parallel p(\theta \mid \mathcal{D})).
\end{equation}
When considering both $p(\theta)$ and $q(\theta)$ as Gaussian distributions with diagonal covariances:
\begin{equation} \label{eq:GaussianDistri}
p(\theta):=\mathcal{N}(\theta \mid \mathbf{0},\mathbf{I}/ \delta), \ \ q(\theta):=\mathcal{N}(\theta \mid \mu, \textup{diag}(\sigma ^2)),
\end{equation}
where $\delta$ is a known precision parameter with $\delta>0$, the mean $\mu$ and deviation $\sigma ^2$ of $q$ can be estimated by minimizing the negative of evidence lower bound (ELBO) \cite{blei2017variational}:
\begin{equation} \label{eq:ELBO}
\resizebox{.9\linewidth}{!}{$
    \displaystyle
\begin{aligned}
\mathcal{L}(\mu ,\sigma):&=-\sum_{i=1}^{N}\mathbb{E}_q \left [ \textup{log}\ p(\mathcal{D}_i \mid \theta) \right ]+ \textup{KL}(q(\theta)\parallel p(\theta))\\
&=-\mathbb{E}_q \sum_{i=1}^{N}  \textup{log}\ p(\mathcal{D}_i \mid \theta)+ \mathbb{E}_q \left [ \textup{log}\ \frac{q(\theta)}{p(\theta)} \right ] \\
\end{aligned}
$}
\end{equation}

A straightforward approach is using the stochastic gradient descent to learn $\mu$ and $\sigma ^2$ along with minimizing $\mathcal{L}$, called as the Bayes by Backprob (BBB) \cite{blundell2015weight}:
\begin{equation} \label{eq:BBB}
\mu_{t+1}=\mu_t-\varsigma_t \hat{\nabla}_\mu \mathcal{L}_t, \ \ \sigma_{t+1}=\sigma_t-\varphi_t \hat{\nabla}_\sigma \mathcal{L}_t,
\end{equation}
where $\varsigma_t$ and $\varphi_t$ are the learning rates, and $\hat{\nabla}_\mu\mathcal{L}_t$ and $\hat{\nabla}_\sigma\mathcal{L}_t$ are the unbiased stochastic gradient estimates of $\mathcal{L}$ at $\mu_t$ and $\sigma_t$. However, VI remains to be impractical for learning large deep networks. The obvious issue is that VI introduces more parameters to learn, as it needs to replace all neural networks weights with random variables and simultaneously optimize two vectors $\mu$ and $\sigma$ to estimate the distribution of $\theta$, so the memory requirement is also doubled, leading a lot of modifications when fitting existing differentiable NAS codebases with the variational inference.


\subsection{Training Free Proxies for NAS}
\label{sec2.3}

Training Free NAS tries to identify promising architectures at initialization without incurring training. Mellor \etal~\cite{mellor2020neural} empirically find that the correlation between sample-wise input-output Jacobian can indicate the architecture’s test performance, and propose using the Jacobian to score a set of randomly sampled models with randomly initialized weights, which greedily chooses the model with the highest score. TE-NAS \cite{chen2021neural} utilizes the spectrum of NTKs and the number of linear regions to analyzing the trainability and expressivity of architectures. Rather than evaluating the whole architecture, TE-NAS uses the perturbation-based architecture selection as \cite{Rethinking2021}, to measure the importance of each operation for the supernet prune. 

Zero-cost NAS \cite{abdelfattah2021zero} extends the saliency metrics in the network pruning at initialization to score an architecture, through summing scores of all parameters $\theta$ in the architecture. There are three popular saliency metrics, SNIP~\cite{lee2018snip}, GraSP~\cite{wang2019picking}, and Synflow~\cite{tanaka2020pruning}:
\begin{equation} \label{eq:zero_proxies}
\resizebox{.9\linewidth}{!}{$
    \displaystyle
\mathcal{S}_{snip}(\theta)=\left | \frac{\partial \mathcal{L}}{\partial \theta}\odot \theta \right |,\ 
\mathcal{S}_{grasp}(-\theta)= -(H \frac{\partial \mathcal{L}}{\partial \theta})\odot \theta,\ 
\mathcal{S}_\textup{SF}(\theta)= \frac{\partial \mathcal{R}_\textup{SF}}{\partial \theta}\odot \theta,
$}
\end{equation}
where $\mathcal{L}$ is the common loss based on initialized weights, $H$ is the Hessian matrix, and $\mathcal{R}_\textup{SF}$ is defined as $\mathcal{R}_{\textup{SF}} = \mathbf{1}^T \left ( \prod_{l=1}^{L} \left | \theta^{[l]} \right | \right ) \mathbf{1}$ that makes SynFlow data-agnostic. Since these scores can be obtained without any training, zero-cost NAS utilizes these zero-cost proxies to assist NAS by \textit{warmup} different search algorithms, e.g., initializing population or controller for aging evolution NAS and RL based NAS, respectively. Different from zero-cost NAS that leverages proxies before the search, we utilize these zero-cost proxies for the architecture selection after search, to select more competitive architectures from the optimized distributions.

\section{The Proposed Method: BaLeNAS}

\label{sec3}



\subsection{Formulating NAS as Distribution Learning}

Differentiable NAS normally considers the architecture parameters $\alpha_\theta$ as learnable parameters and directly conducts optimization in this space. Most previous differentiable NAS methods first optimize the architecture parameters based on the gradient of the performance, then update the supernet weights based on the updated architecture parameters. Since architectures with updated supernet weights are supposed to have higher performance, architectures with better performance in the early stage have a higher probability of being selected for the supernet training. The supernet training again improves these architectures' performance. This is to say, directly optimizing $\alpha_\theta$ without exploration easily entails the \textit{rich-get-richer  problem} \cite{li2019improving,zhang2020one}, leading to suboptimal paths in the search space that converges faster at the beginning but plateaued quickly \cite{shu2019understanding,chen2020drnas}. In contrast, formulating the differentiable NAS as a distribution learning problem by relaxing architecture parameters can naturally introduce \textbf{stochasticity} and encourage \textbf{exploration} to resolve this problem \cite{chen2020drnas,chen2020stabilizing}.

In this paper, we formulate the architecture search as a distribution learning problem, that for the first time consider the more general Gaussian distributions for the architecture parameters to optimize the posterior distribution $p(\alpha_\theta \mid \mathcal{D})$ rather than $\alpha_\theta$. Considering both $p(\theta)$ and $q(\theta)$ as Gaussian distributions as Eq.\eqref{eq:GaussianDistri}, the bilevel optimization problem in Eq.\eqref{eq:bilevel} could be reformulated as the distribution learning based NAS:
\begin{equation} \label{eq:distri_lear}
\begin{aligned}
&\underset{\mu,\sigma }{\mathtt{min}} \ \mathbb{E}_{q(\alpha_\theta \mid \mu,\sigma )} \mathcal{L}_{\mathtt{val}}(w^*(\alpha_\theta),\alpha_\theta), \\
&\textup{s.t.}\ w^*(\alpha_\theta)=\underset{w}{\mathtt{argmin}} \mathcal{L}_{\mathtt{train}}(w(\alpha_\theta),\alpha_\theta),
\end{aligned}
\end{equation}
where $\mu$ and $\sigma$ are the two learnable parameters for the distribution $q(\alpha_\theta \mid \mu,\sigma ):=\mathcal{N}(\alpha_\theta \mid \mu, \textup{diag}(\sigma ^2))$. Considering the variational inference and Bayesian deep learning, based on Eq.\eqref{eq:minKL}-\eqref{eq:ELBO}, the loss function for the outer-loop architecture distribution optimization problem could be defined as:
\begin{equation} \label{eq:ELBO_darts}
\resizebox{.9\linewidth}{!}{$
    \displaystyle
\mathbb{E}_{q} \left [\mathcal{L}_{\mathtt{val}}\right ]:=-\mathbb{E}_q \sum_{i=1}^{N}  \textup{log}\ p(\mathcal{D}_i \mid \alpha_\theta)+ \mathbb{E}_q \left [ \textup{log}\ \frac{q(\alpha_\theta)}{p(\alpha_\theta)} \right ].
$}
\end{equation}
%
Since the architecture parameters $\alpha_\theta$ are random variables sampled from the Gaussian distribution $q(\alpha_\theta \mid \mu,\sigma )$, the distribution learning-based method naturally encourages exploration during the architecture search.

\subsection{Natural-Gradient VI for NAS}
\label{sec3.2}

As describe in Sec.\ref{sec2.2}, the traditional variational inference has double memory requirement and needs to re-design the object function, making it difficult to fit with the differentiable NAS. Thus, this paper considers natural-gradient variational inference (NGVI) methods \cite{khan2018fast,osawa2019practical} to optimize the architecture distribution $p(\alpha_\theta \mid \mathcal{D})$ in a natural parameter space, which requires the same number of parameters as the traditional learning method. By leveraging NGVI, the architecture parameter distribution could be learned by only updating a natural parameter $\lambda$ during the search.

NGVI parameterizes the distribution $q(\alpha_\theta)$ with a natural parameter $\lambda$, considering $q(\alpha_\theta \mid \lambda)$ in a class of minimal exponential family with natural parameter $\lambda$ \cite{khan2017conjugate}:
\begin{equation} \label{eq:naturalPara}
q(\alpha_\theta \mid \lambda) :=h(\alpha_\theta)\textup{exp}\left [ \lambda^T\phi(\alpha_\theta)-A(\lambda) \right ],
\end{equation}
where $h(\alpha_\theta)$ is the base measure, $\phi(\alpha_\theta)$ is a vector containing sufficient statistics, and $A(\lambda)$ is the log-partition function.

When $h(\alpha_\theta)\equiv 1$, the distribution $q(\alpha_\theta \mid \lambda)$ could be learned by only updating $\lambda$ during the training \cite{khan2018fast,khan2020}, and $\lambda$ could be learned in the natural-parameter space by:
\begin{equation} \label{eq:natureupdate}
\lambda_{t+1}= (1-\rho_t)\lambda_t-\rho_t \nabla_{\mu}\mathbb{E}_{q_t}\left [ \bar{\ell}(\alpha_\theta) \right ],
\end{equation}
where $\rho_t$ is the learning rate, $\bar{\ell}$ is in the form of Eq.\eqref{eq:erm}, and the derivative $\nabla_{\mu}\mathbb{E}_{q_t(\alpha_\theta)}\left [ \bar{\ell}(\alpha_\theta) \right ]$ is taken at $\mu=\mu_t$ which is the expectation parameter with Markov Chain Monte Carlo (MCMC) sampling. And $q_t$ is the $q(\alpha_\theta \mid \lambda)$ parameterized by $\lambda_{t}$, $\mu=\mu(\lambda)$ is the expectation parameter of $q(\alpha_\theta \mid \lambda)$. This is also called as the Bayesian learning rule \cite{khan2020}.


When we consider Gaussin mean-field VI that $p(\alpha_\theta)$ and $q(\alpha_\theta)$ are in the form of Eq.\eqref{eq:GaussianDistri}, the Variational Online-Newton (VON) method proposed by Khan et. al. \cite{khan2018fast} shows that the NGVI update could be written with the following update:
\begin{equation} \label{eq:muupdate}
\mu_{t+1}=\mu_t-\beta_t(\hat{\textbf{g}}(\theta_t)+\tilde{\delta}\mu_t)/(\textbf{s}_{t+1}+\tilde{\delta}),
\end{equation}
\begin{equation} \label{eq:sigmaupdate}
\textbf{s}_{t+1}=(1-\beta_t)\textbf{s}_t+\beta_t\ \textup{diag}[\hat{\nabla}^2\bar{\ell}(\theta_t)],
\end{equation}
where $\beta_t$ is the learning rate, $\theta_t\sim \mathcal{N}({\alpha_\theta} \mid \mu_t,\sigma_t^2)$ with $\sigma_t^2=1/[N(\textup{s}_t+\tilde{\delta})]$ and $\tilde{\delta}=\delta/N$. $\hat{\textbf{g}}$ is the stochastic estimate with respect to $q$ through MCMC sampling that, $\hat{\mathbf{g}}(\theta_t)=\frac{1}{M}\sum_{i\in \mathcal{M}}\nabla_{\alpha_\theta} \bar{\ell}_i(\alpha_\theta)$, and the minibatch $\mathcal{M}$ contains $M$ samples. More details are in \cite{khan2018fast}. Variational RMSprop (Vprop) \cite{khan2018fast} further uses gradient magnitude (GM) \cite{bottou2018optimization} approximation to reformulate Eq.\eqref{eq:sigmaupdate} as:
\begin{equation} \label{eq:sigmaupdate2}
\textbf{s}_{t+1}=(1-\beta_t)\textbf{s}_t+\beta_t[\hat{\textbf{g}}(\theta_t)\circ \hat{\textbf{g}}(\theta_t)],
\end{equation}
with $\hat{\nabla}^2_{j,j}\bar{\ell}(\theta_t)\approx \left [  \frac{1}{M}\sum_{i\in \mathcal{M}_t}g_i(\alpha_\theta^j)\right ]^2 =[\hat{g}(\theta_t^j)]^2$ \cite{bottou2018optimization}. The most important benefit of VON and Vprop is that they only need to calculate one parameter's gradient to update posterior distribution. In this way, this learning paradigm requires the same number of parameters as traditional learning methods and easy to fit with existing codebases.


We implement the proposed BaLeNAS based on the DARTS \cite{liu2018darts} framework, the most popular differentiable NAS baseline. Similar to DARTS, BaLeNAS also considers an Adam-like optimizer for the architecture optimization, updating the natural parameter $\lambda$ of $p(\theta \mid \mathcal{D})$ as:
\begin{equation} \label{eq:vadamupdate}
\lambda_{t+1}= \lambda_t-\rho_t \nabla_{\lambda}\mathcal{L}_t+\gamma_t(\lambda_t-\lambda_{t-1}),
\end{equation}
where the last term is the momentum. Based on the Vprop in Eq.\eqref{eq:muupdate} and \eqref{eq:sigmaupdate2}, the update of $\mu$ and $\sigma$ for the Adam-like optimizer with NGVI, also called as Variational Adam (VAdam), could be defined as following:
\begin{equation} \label{eq:vadammu}
\begin{aligned}
\mu_{t+1}=&\mu_t-\beta_t(\hat{\textbf{g}}(\theta_t)+\tilde{\delta}\mu_t) \circ \frac{1}{(\textbf{s}_{t+1}+\tilde{\delta})}\\
&+\gamma_t\left \lfloor \frac{\mathbf{s}_t+\tilde{\delta}}{\mathbf{s}_{t+1}+\tilde{\delta}} \right \rfloor \circ (\mu_{t}-\mu_{t-1}),
\end{aligned}
\end{equation}
\begin{equation} \label{eq:vadamsigma}
\textbf{s}_{t+1}=(1-\beta_t)\textbf{s}_t+\beta_t[\hat{\textbf{g}}(\theta_t)\circ \hat{\textbf{g}}(\theta_t)].
\end{equation}
where ``$\circ$" stands for element-wise product, $\theta_t\sim \mathcal{N}({\alpha_\theta} \mid \mu_t,\sigma_t^2)$ with $\sigma_t^2=1/[N(\textup{s}_t+\tilde{\delta})]$. As pointed out in Sec. \ref{sec2.2} and shown in Eq.\eqref{eq:vadammu} and Eq.\eqref{eq:vadamsigma}, the distribution  $q(\alpha_\theta)=\mathcal{N}(\alpha_\theta \mid \mu, \sigma^2)$ is now optimized, needing to calculate the gradient of only one parameter.

\vspace{1mm}
\noindent\textbf{Implicit Regularization from MCMC Sampling:}
Several recent works \cite{zela2019understanding,chen2020stabilizing,chen2020drnas} empirically and theoretically show that the performance of differentiable NAS is highly related to the norm of $\mathcal{H}$, the Hessian matrix of $\mathcal{L}_{val}(w^*,\alpha_{\theta})$, and keeping this norm in a low level plays a key role in robustifying differentiable NAS. As described before, we know the loss $\mathbb{E}_{q_t(\alpha_\theta)}\left [ \bar{\ell}(\alpha_\theta) \right ]$ of architecture optimization in BaLeNAS is calculated based on MCMC sampling, showing the naturality of enhancing exploration. Besides, $\mathbb{E}_{q_t(\alpha_\theta)}\left [ \bar{\ell}(\alpha_\theta) \right ]$ also has the naturality to enhance the stability in differentiable NAS as SDARTS \cite{chen2020stabilizing}. When conducting the Taylor expansion, the loss function for the architecture parameters update $\mathbb{E}_{q_t(\alpha_\theta)}\left [ \bar{\ell}(\alpha_\theta) \right ]$ could be described as: 
\begin{equation} \label{eq:stabilizing}
\resizebox{.85\linewidth}{!}{$
    \displaystyle
\begin{aligned}
&\mathbb{E}_{q_t(\alpha_\theta)}\left [ \bar{\ell}(\alpha_\theta) \right ]\\
=&\mathbb{E}_{q(\alpha_\theta \mid \mu,\sigma )} \mathcal{L}_{val}(w,\alpha_\theta)= \mathbb{E}_{\epsilon\sim \mathcal{N}(0,\sigma^2)}\mathcal{L}_{val}(w,\mu+\epsilon)\\
=&\mathbb{E}_{\epsilon\sim \mathcal{N}(0,\sigma^2)}[ \mathcal{L}_{val}(w,\mu)+\triangledown_{\mu}\mathcal{L}_{val}(w,\mu)^T\epsilon+\frac{1}{2} \epsilon^T \mathcal{H}\epsilon ]\\
=&\mathbb{E}_{\epsilon\sim \mathcal{N}(0,\sigma^2)}\left [\mathcal{L}_{val}(w,\mu)+\frac{1}{2}\epsilon^T\mathcal{H}\epsilon\right ]\\
=&\mathcal{L}_{val}(w,\mu)+\frac{\sigma^2}{2}\text{Tr}\left \{ \mathcal{H} \right \},
\end{aligned}
$}
\end{equation}
where the line 4 in Eq.\eqref{eq:stabilizing} is obtained since $\mathbb{E}_{\epsilon\sim \mathcal{N}(0,\sigma^2)}[\triangledown_{\mu} \mathcal{L}_{val}(w,\alpha_{\theta})^T\epsilon]=\mathbb{E}_{\epsilon\sim \mathcal{N}(0,\sigma^2)}[\epsilon]*\triangledown_{\mu} \mathcal{L}_{val}(w,\alpha_{\theta})=0$, as $\epsilon\sim\mathcal{N}(0,\sigma^2)$ is a Gaussian distribution with zero mean, and $\mathbb{E}(\epsilon^2)=\sigma^2$. $\mu$ is the expectation parameter of $q(\alpha_\theta \mid \mu,\sigma^2 )$, and $\mathcal{H}$ is the Hessian matrix of $\mathcal{L}_{val}(w,\mu)$. We can find the loss function that could implicitly control the trace norm of $\mathcal{H}$ similar as \cite{chen2020stabilizing,chen2020drnas}, helping \textbf{stabilizing} differentiable NAS.

\begin{algorithm}[t]
\caption{BaLeNAS}
\label{alg:algorithm1}
Initialize a supernet with supernet weights $w$ and architecture parameters $\alpha_\theta$
\begin{algorithmic}[2]
\WHILE{\textit{not converged}}
\STATE Update {\color{red} $\mu$ and $\sigma^2$ for $q({\alpha_\theta} \mid \mu,\sigma^2)$ based on Eq.\eqref{eq:vadammu} and Eq.\eqref{eq:vadamsigma}, with VAdam optimizer.}
\STATE Update supernet weights $w$ based on cross-entropy loss with the common SGD optimizer.
\ENDWHILE
\STATE Obtain discrete architecture $\alpha^*$ through \textit{argmax} on $\mu$; or {\color{red}sample a set of $\alpha_\theta$ from $q({\alpha_\theta^*} \mid \mu,\sigma^2)$, and utilize the training free proxies for selection}.
\end{algorithmic}
\end{algorithm}

\subsection{Architecture Selecting from the Distribution}
After the optimization of BaLeNAS, we learns an optimized Gaussian distribution for the architecture parameters $q({\alpha_\theta^*} \mid \mu,\sigma^2)$, which is used to get the optimal architecture $\alpha^*$. In this paper, we consider two methods to get the discrete architecture $\alpha^*$. The first one is a simple and direct method, which utilizes the expectation of $\alpha_\theta^*$ to select the best operation for each edge through the \textit{argmax} as DARTS, where the expectation term is simply the mean $\mu$ \cite{chen2020drnas}. However, as we described in Sec. \ref{sec2.1}, this method may result in instability and incongruence. The second one is more general, which samples a set of $\alpha$ from the distribution $q({\alpha_\theta^*} \mid \mu,\sigma^2)$ for architecture selection. However, in the neural architecture search, evaluating a set of architectures will incur unaffordable computational costs. In this paper, instead of utilizing training-free proxies to assist NAS by \textit{warmup} before search as \cite{abdelfattah2021zero}, we leverage these proxies, including SNIP~\cite{lee2018snip}, GraSP~\cite{wang2019picking}, and Synflow~\cite{tanaka2020pruning}, to score the sampled architectures for selection after search.

Algorithm \ref{alg:algorithm1} gives a simple implementation of BaLeNAS, where only the red part is different from DARTS. As shown, in our BaLeNAS, only architecture parameter optimization is different from DARTS which uses the VAdam optimizer, making it easy to be implemented. Furthermore, as most existing differentiable NAS methods are built based on DARTS codebase, our BaLeNAS is also comfortable to be adapted to them with minimal modifications.


\begin{table*}[ht]
\centering
\caption{Comparison results with state-of-the-art NAS approaches on NAS-Bench-201. 
}
\small
\begin{tabular}
{|l|c|c|c|c|c|c|c|c|}
\hline

\multirow{2}*{Method}&\multicolumn{2}{c|}{CIFAR-10}&\multicolumn{2}{c|}{CIFAR-100}&\multicolumn{2}{c|}{ImageNet-16-120}\\
~&\multicolumn{1}{c}{Valid(\%)}&\multicolumn{1}{c|}{Test(\%)}&\multicolumn{1}{c}{Valid(\%)}&\multicolumn{1}{c|}{Test(\%) }&\multicolumn{1}{c}{Valid(\%) }&\multicolumn{1}{c|}{Test(\%)}\\
\hline
\hline
Random baseline&83.20$\pm$13.28&86.61$\pm$13.46&60.70$\pm$12.55&60.83$\pm$12.58&33.34$\pm$9.39&33.13$\pm$9.66\\
ENAS \cite{pham2018efficient}&37.51$\pm$3.19&53.89$\pm$0.58&13.37$\pm$2.35&13.96$\pm$2.33&15.06$\pm$1.95&14.84$\pm$2.10\\
RandomNAS \cite{li2019random}&85.63$\pm$0.44&88.58$\pm$0.21&60.99$\pm$2.79&61.45$\pm$2.24&31.63$\pm$2.15&31.37$\pm$2.51\\
SETN \cite{dong2019one}&84.04$\pm$0.28&87.64$\pm$0.00&58.86$\pm$0.06&59.05$\pm$0.24&33.06$\pm$0.02&32.52$\pm$0.21\\
GDAS \cite{GDAS}&90.00$\pm$0.21&93.51$\pm$0.13&71.14$\pm$0.27&70.61$\pm$0.26&41.70$\pm$1.26&41.84$\pm$0.90\\
DrNAS \cite{chen2020drnas}&91.55$\pm$0.00&94.36$\pm$0.00&73.49$\pm$0.00&73.51$\pm$0.00&46.37$\pm$0.00&46.34$\pm$0.00\\
DARTS (1st) \cite{liu2018darts}&39.77$\pm$0.00&54.30$\pm$0.00&15.03$\pm$0.00&15.61$\pm$0.00&16.43$\pm$0.00&16.32$\pm$0.00\\
DARTS (2nd) \cite{liu2018darts}&39.77$\pm$0.00&54.30$\pm$0.00&15.03$\pm$0.00&15.61$\pm$0.00&16.43$\pm$0.00&16.32$\pm$0.00\\
Zero-cost NAS \cite{abdelfattah2021zero}&90.19$\pm$0.66&93.45$\pm$0.28&70.55$\pm$1.61&70.73$\pm$1.36&43.24$\pm$2.52&43.64$\pm$2.42\\
\hline
BaLeNAS (1st) &91.03$\pm$0.15&93.62$\pm$0.12&70.88$\pm$0.60&70.98$\pm$0.41&45.19$\pm$0.75&45.25$\pm$0.86\\
BaLeNAS (2nd) &91.32$\pm$0.09&94.02$\pm$0.14&71.53$\pm$0.08&71.93$\pm$0.27&45.39$\pm$0.17&45.48$\pm$0.39\\
BaLeNAS-TF &91.52$\pm$0.04&94.33$\pm$0.03&72.67$\pm$0.41&72.95$\pm$0.28&46.14$\pm$0.23&46.54$\pm$0.36\\
\hline
\textbf{optimal}&91.61&94.37&74.49&73.51&46.77&47.31\\
\hline
\end{tabular}
\flushleft{The best single run of BaLeNAS-TF achieves \textbf{94.37\%}, \textbf{73.22\%}, and \textbf{46.71\%} test accuracy on three datasets, respectively. Our BaLeNAS-TF considers the Synflow based proxy for architecture selection in this experiment.}
\label{tab:nasbench201}
\vspace{-1em}
\end{table*}

\section{Experiments and Results}
In this section, we consider three different search spaces to analyze the proposed BaLeNAS framework. The first two are NAS benchmark datasets, NAS-Bench-201 \cite{BENCH102} and NAS-Bench-1shot1 \cite{zela2020nasbench1shot1}. The ground-truth for all candidate architectures in the two benchmark datasets is known. The NAS methods could be evaluated without retraining the searched architectures based on these benchmark datasets, thus greatly relieving the computational burden. The third one is the commonly-used CNN search space in DARTS \cite{liu2018darts}. We first analyze our proposed BaLeNAS in the two benchmark datasets, then compare BaLeNAS with state-of-the-art NAS methods in the DARTS search space. 


\vspace{-1mm}
\subsection{Experiments on Benchmark Datasets}
The NAS-Bench-201 \cite{BENCH102} has a unified cell-based search space, where the cell structure is densely-connected, containing four nodes with five candidate operations applied on each node, resulting in 15,625 architectures. NAS-Bench-201 reports the CIFAR-10, CIFAR-100, and Imagenet performance for all architecture in this search space. The NAS-Bench-1shot1 \cite{zela2020nasbench1shot1} is built from the NAS-Bench-101 benchmark dataset \cite{ying2019bench}, through dividing all architectures in NAS-Bench-101 into 3 different unified cell-based search spaces, containing 6,240, 29,160, and 363,648 architectures, respectively, and the CIFAR-10 performance for all architectures are reported. The architectures in each search space have the same number of nodes and connections, making the differentiable NAS could be directly applied to each space.

\begin{figure}[t]
\centering
 \subfloat[First order approximation]{
  \begin{minipage}{4cm}
      \includegraphics[width=4.2cm,height=2.9cm]{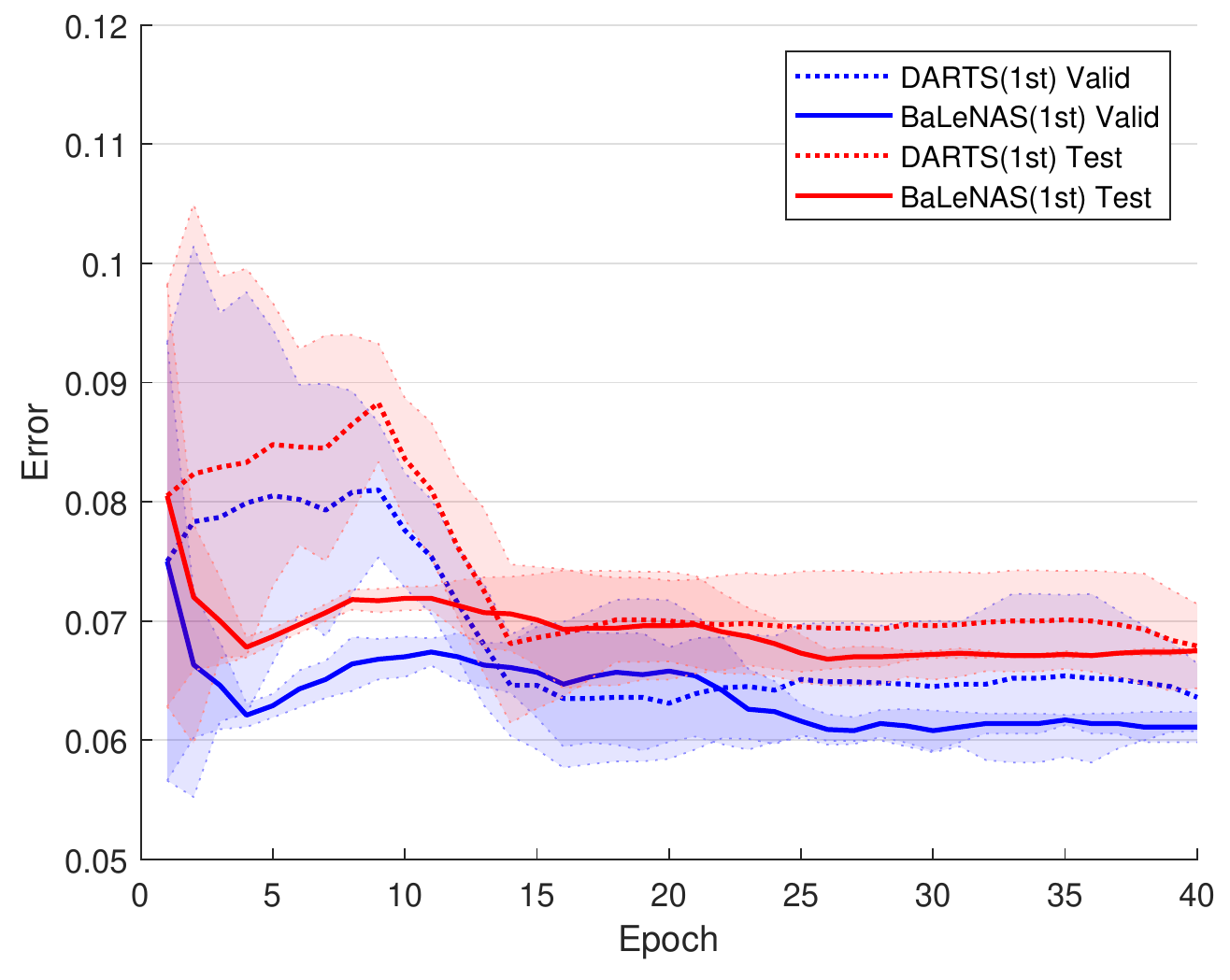}
  \end{minipage}
 }
  \subfloat[Second order approximation]{
  \begin{minipage}{4cm}
       \includegraphics[width=4.2cm,height=2.9cm]{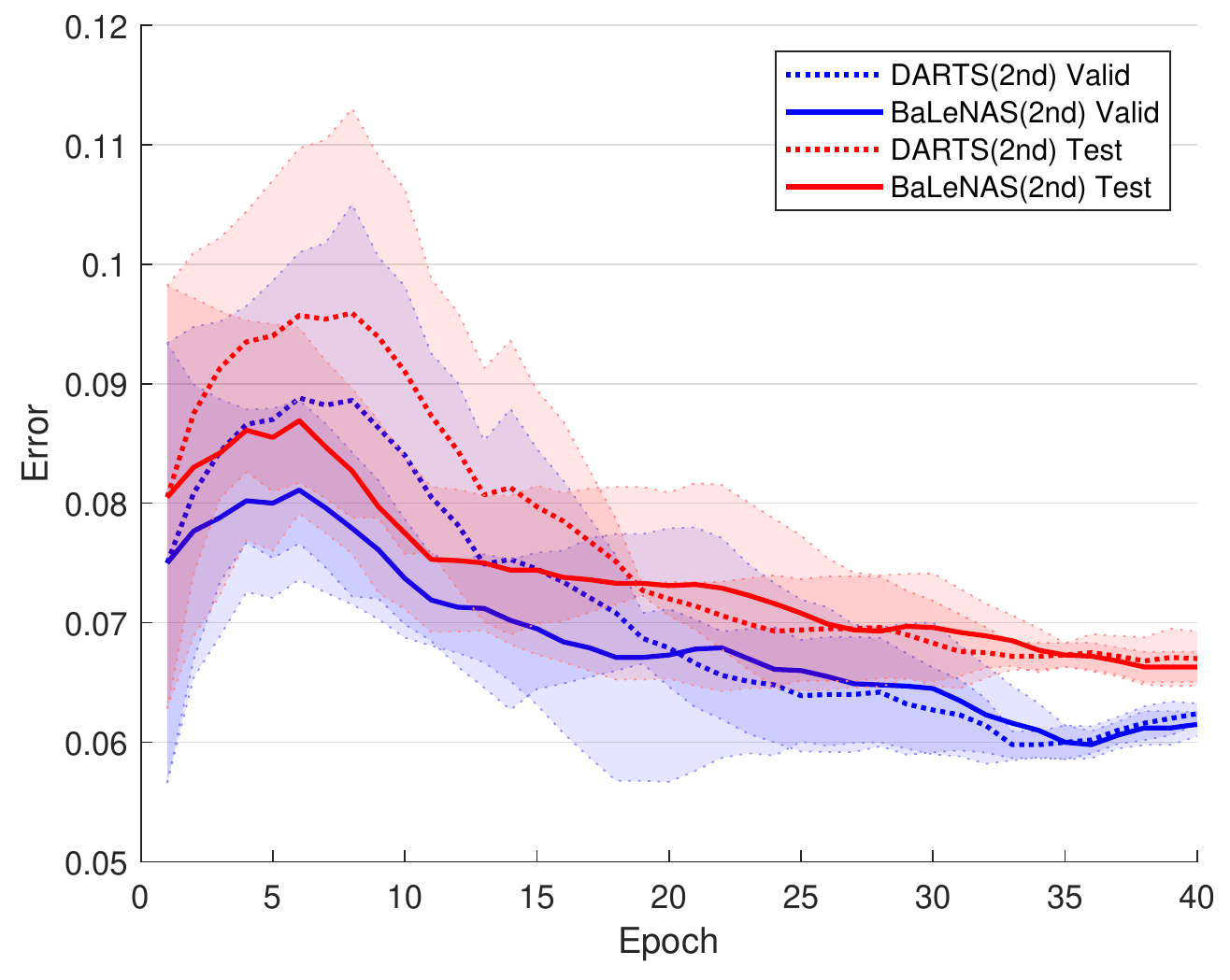}
  \end{minipage} 
  }
  
 \caption{Validation and test error of BaLeNAS and DARTS on the search space 3 of NAS-Bench-1shot1. }
 \label{fig:compa_nas1shot1}
 \vspace{-2mm}
\end{figure}

\vspace{-2mm} 
\subsubsection{Reproducible Comparison on NAS Benchmarks}
Table \ref{tab:nasbench201} summarizes the performance of BaleNAS on NAS-Bench-201 compared with differentiable NAS baselines, where the statistical results are obtained from 4 independent search experiments with four different \textit{random seeds}. In our BaLeNAS, we consider the expectation of $\alpha_\theta$ with \textit{argmax} to get the valid architecture, while BaLeNAS-TF consider the training-free proxies for the architecture selection, with the sample size is set as 100. As shown in Table \ref{tab:nasbench201}, BaLeNAS achieves the best results on the NAS-Bench-201 benchmark and greatly outperforms other baselines on all three datasets. As described in Sec. \ref{sec3}, BaLeNAS is built based on the DARTS framework, with only modeling the architecture parameters into distributions and introducing Bayesian learning rule for optimization. As shown in Table \ref{tab:nasbench201}, BaLeNAS with first and second-order approximations both outperform DARTS by large margins, verifying the effectiveness of our method. More interesting, combining with the training-free proxies, BaLeNAS-TF can achieve better results, showing that apart from \textit{warmup}, these proxies could also assist differentiable NAS at architecture selection. The best single run of our BaLeNAS-TF achieves \textbf{94.37\%}, \textbf{73.22\%}, and \textbf{46.71\%} test accuracy on three datasets, respectively, which are state-of-the-art on this benchmark dataset.


\begin{table}
\footnotesize
\caption{Ablation study on the sample size.}
\begin{tabular}{lccc}
\toprule
\multirow{2}*{Method (size)}&\multicolumn{3}{c}{Test Accuracy}\\
~&CIFAR-10&CIFAR-100&ImageNet\\
\midrule
Zero-cost NAS(10)&92.12$\pm$1.25&68.1$\pm$2.49&40.07$\pm$1.86\\
Zero-cost NAS(50)&92.52$\pm$0.05&70.27$\pm$0.25&42.92$\pm$0.95\\
Zero-cost NAS(100)&93.45$\pm$0.16&69.87$\pm$0.35&44.43$\pm$0.75\\
BaLeNAS-TF(10)&94.08$\pm$0.13&72.55$\pm$0.42&45.82$\pm$0.30\\
BaLeNAS-TF(50)&\textbf{94.33$\pm$0.03}&\textbf{72.95$\pm$0.28}&\textbf{46.54$\pm$0.36}\\
BaLeNAS-TF(100)&\textbf{94.33$\pm$0.03}&\textbf{72.95$\pm$0.28}&\textbf{46.54$\pm$0.36}\\
\bottomrule
\end{tabular}
\label{tab:results_ablation_proxy}
\vspace{-2em}
\end{table}

\begin{table*}[htb]
\centering
\caption{Comparison results with state-of-the-art weight-sharing NAS approaches.}

\begin{tabular}{|l|c|c|c|c|c|c|c|c|}
\hline

\multirow{2}*{Method}&\multicolumn{3}{c|}{Test Error (\%)}&\multicolumn{1}{c|}{Param}&\multicolumn{1}{c|}{FLOPs}&\multicolumn{1}{c|}{Search}&\multicolumn{1}{c|}{Architecture}\\
~&\multicolumn{1}{c}{CIFAR-10}&\multicolumn{1}{c}{CIFAR-100}&\multicolumn{1}{c|}{ImageNet}&{(M)}&{(M)}&{Cost}&{Optimization}\\
\hline\hline
RandomNAS \cite{li2019random}&2.85$\pm$0.08&17.63 &27.1&4.3&595&2.7&random\\
SNAS \cite{xie2018snas}&2.85$\pm$0.02&20.09 &27.3 / 9.2&2.8&467&1.5&gradient\\
BayesNAS \cite{zhou2019bayesnas}&2.81$\pm$0.04&-&26.5 / 8.9&3.40&-&0.2&gradient\\
MdeNAS \cite{zheng2019multinomial}&2.55&17.61&25.5 / 7.9&3.61&500&0.16&gradient\\
GDAS \cite{GDAS}&2.93&18.38&26.0 /  8.5&3.4&538&0.21&gradient\\
XNAS \cite{nayman2019xnas}&2.57$\pm$0.09&16.34&24.7 / 7.5&3.7&590&0.3&gradient\\
PDARTS \cite{chen2019progressive}&2.50&16.63& 24.4 / 7.4&3.4&543&0.3&gradient\\
PC-DARTS \cite{xu2019pcdarts}&2.57$\pm$0.07&17.11&25.1 / 7.8&3.6&571&0.3&gradient\\
DrNAS \cite{chen2020drnas}&2.54$\pm$0.03&16.30&\textbf{24.2 / 7.3}&4.0&644&0.4&gradient\\
DARTS+ \cite{liang2019darts+}&2.50$\pm$0.11& 16.28&-&3.7&-&0.4&gradient\\
DARTS (1st) \cite{liu2018darts}&2.94&-&-&2.9&505&1.5&gradient\\
DARTS (2nd) \cite{liu2018darts}&2.76$\pm$0.09&17.54&26.9 / 8.7&3.4&530&4&gradient\\
\hline\hline
BaLeNAS&2.50$\pm$0.07&16.84&25.0 / 7.7&3.82&593&0.6&gradient\\
BaLeNAS-TF&\textbf{2.43$\pm$0.08} & \textbf{15.72}&\textbf{24.2 / 7.3}&3.86&597&0.6&gradient\\
\hline
\end{tabular}

\label{tab:results_CIFAR}
\end{table*}

We also conduct a comparison study on the NAS-Bench-1shot1 dataset to further verify the effectiveness of our BaLeNAS which reformulates architecture search as a distribution learning problem. We have compared BaLeNAS with the baseline DARTS on the three search spaces of NAS-Bench-1shot1 with tracking the validation and test performance of the search architectures in every iteration. As shown in Fig. \ref{fig:compa_nas1shot1}, our BaLeNAS, without training-free proxies based architecture selection, generally outperforms DARTS during the architecture search in terms of validation and test error in the most complicated search space 3, both with first and second-order approximation. More specifically, our BaLeNAS significantly outperforms the baseline in the early stage, demonstrating our BaLeNAS could quickly find the superior architectures and is more stable. The results on both NAS-Bench-201 and NAS-Bench-1shot1 verify that, by formulating the architecture search as a distribution learning problem and introducing the Bayesian learning rule to optimize the posterior distribution, BaLeNAS can relieve the instability and naturally enhance exploration to avoid local optimum for differentiable NAS.

\vspace{-1em}
\subsubsection{Ablation Study on the Architecture Selection}

As described, our BaLeNAS-TF samples several architectures from the optimized distribution and leverages the training-free proxies for architecture selection, rather than simply applying \textit{argmax} on the mean. In this subsection, we conduct ablation study to investigate the benefits of our training-free based architecture selection. We considered 3 different training-free proxies as described in Sec.~\ref{sec2.3}, including SNIP, GraSP, and Synflow. We find that Synflow is the most reliable proxies in the architecture selection, as it achieves better performance than the remaining two proxies for both zero-cost NAS and BaLeNAS, and also consistently enhances the performance with the increase of sample size. More detailed comparison can be found in the Appendix. Zero-cost NAS \cite{abdelfattah2021zero} randomly generates samples and calculates the scores based on the proxies for architecture selection, while our BaLeNAS-TF generates samples based on the optimized distribution $({\alpha_\theta^*} \mid \mu,\sigma^2)$. 

Table \ref{tab:results_ablation_proxy} compared zero-cost NAS and BaLeNAS-TF with different sample sizes in the architecture selection. As shown, the Synflow proxy can assist NAS as zero-cost NAS with different sample sizes achieve much better results than the Random baseline in Table \ref{tab:nasbench201}, and these proxies also enhance our BaLeNAS, where our BaLeNAS-TF achieve higher accuracy. These results again verified that the architecture selection with train-free proxies can further improve the performance for distribution learning based NAS. More interesting, Table \ref{tab:results_ablation_proxy} also showed that our BaLeNAS-TF outperformed zero-cost NAS by a large margin, suggesting that our BaLeNAS can converge to a competitive distribution.



\subsection{Experiments on DARTS Search Space}
To compare with the state-of-the-art differentiable NAS methods, we applied BaLeNAS to the typical DARTS search space \cite{liu2018darts,GDAS,li2019random} for convolutional architecture search, where all experiment settings are following DARTS \cite{liu2018darts} for fair comparisons as the same as the most recent works. Our BaLeNAS-TF also considers the Synflow proxy in this experiment. The architecture search in DARTS space generally contains three stages: The differentiable NAS first searches for micro-cell structures on CIFAR-10, and then stack more cells to form the full structure for the architecture evaluation. The best-found cell on CIFAR-10 is finally transferred to larger datasets to evaluate its transferability. 
\vspace{-2mm}

\subsubsection{Search Results on CIFAR-10}

The comparison results with the state-of-the-art NAS methods are presented in Table \ref{tab:results_CIFAR}. The best architecture searched by our BaLeNAS-TF achieves a 2.37\% test error on CIFAR-10, which outperforms state-of-the-art NAS methods. We can also see that both BaLeNAS-TF and BaLeNAS outperform DARTS by a large margin, demonstrating the effectiveness of the proposed method. Besides, although BaLeNAS introduced MCMC during architecture optimization, it is still efficient in the sense that the whole architecture search phase in BaLeNAS (2nd) only took 0.6 GPU days.

\subsubsection{Transferability Results Analysis}
\label{sec4.2.2}
Following DARTS experimental setting, the best-searched architectures on CIFAR-10 are then transferred to CIFAR-100 and ImageNet to evaluate the transferability. The comparison results with state-of-the-art differentiable NAS approaches on CIFAR-100 and ImageNet are demonstrated in Table \ref{tab:results_CIFAR}. As shown in Table2, BaLeNAS-TF achieves a 15.72\% test error on the CIFAR-100 dataset, which is a state-of-the-art performance and outperforms peer algorithms by a large margin. On the ImageNet dataset, the best-discovered architecture by our BaLeNAS-TF also achieved a competitive result with 24.2 / 7.3 \% top1 / top5 test error, outperforming or on par with all peer algorithms. 
\vspace{-2mm}

\begin{figure}
\centering
      \includegraphics[width=.8\linewidth]{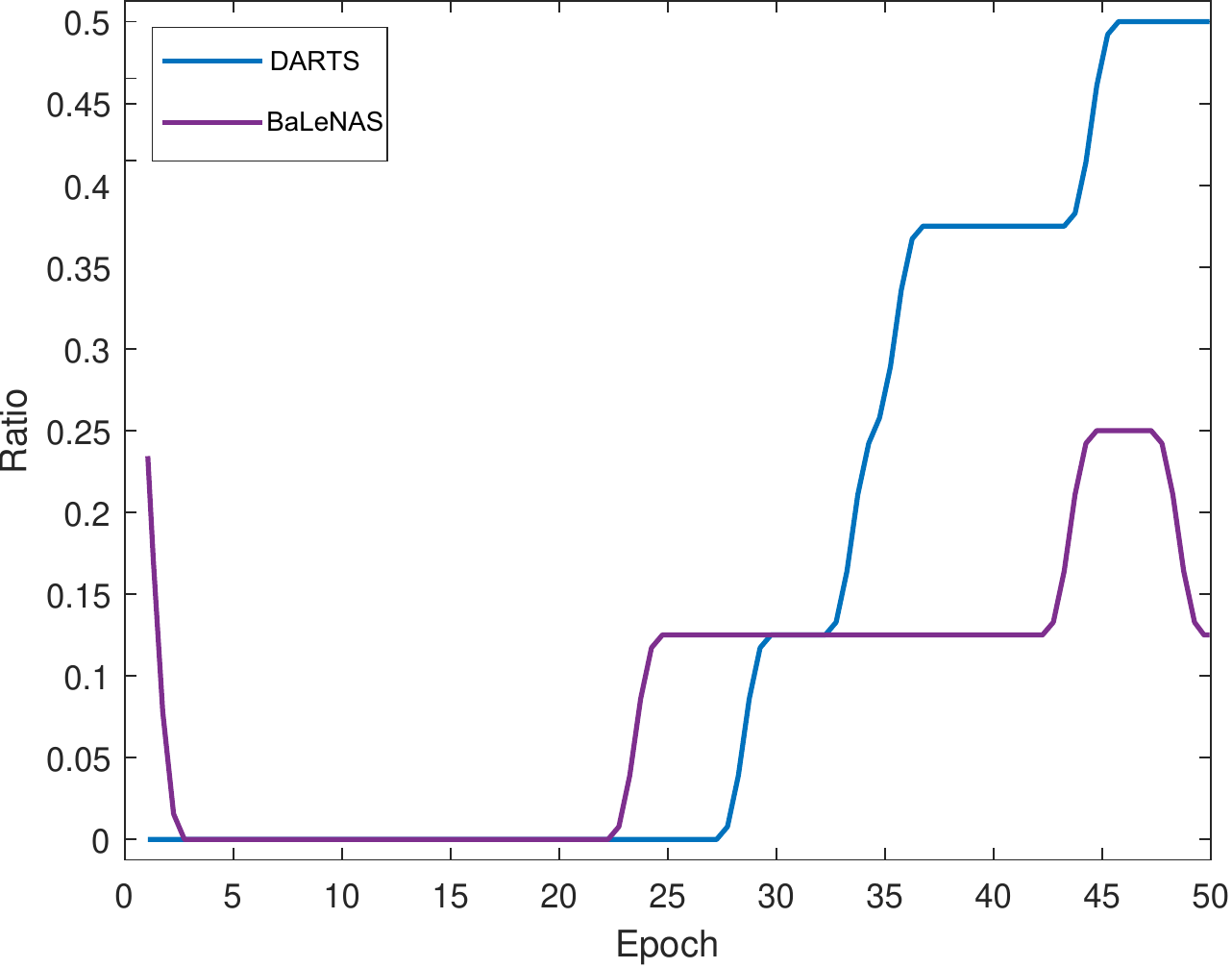}
 \caption{The ratio of skip-connection the searched normal cells during the architecture search in the DARTS space.}
 \label{fig:skip}
 \vspace{-2mm} 
\end{figure}

\subsubsection{Analysis on the Effect of Exploration}
Several recent works \cite{shu2019understanding,chen2020drnas,zhang2020one} point out that directly optimizing architecture parameters without exploration easily entails the rich-gets-richer problem, leading to those architectures that converge faster at the beginning while achieve poor performance at the end of training, e.g. architectures with intensive \textit{skip-connections} \cite{chu2019fairnas,liang2019darts+}. However, when the number of \textit{skip-connections} is larger than 3, the architecture's retraining accuracy is usually extremely low \cite{liang2019darts+,zela2019understanding}. To relieve this issue, BaLeNAS formulates the differentiable neural architecture search as a distribution learning problem, and this experiment verifies how the proposed formulation naturally enhance the exploration to relieve this issue. Fig. \ref{fig:skip} plots the ratio of \textit{skip-connection} in the searched normal cell for BaLeNAS and DARTS (the total number of operations in a cell is 8). As shown, DARTS is likely to select more than 3 \textit{skip-connection} in the normal cell during the search. In contrast, in the proposed BaLeNAS, the number of \textit{skip-connections} is generally less than 2 in the normal cell during the search for BaLeNAS.

\begin{figure}[t]
  \centering
  \includegraphics[width=\linewidth]{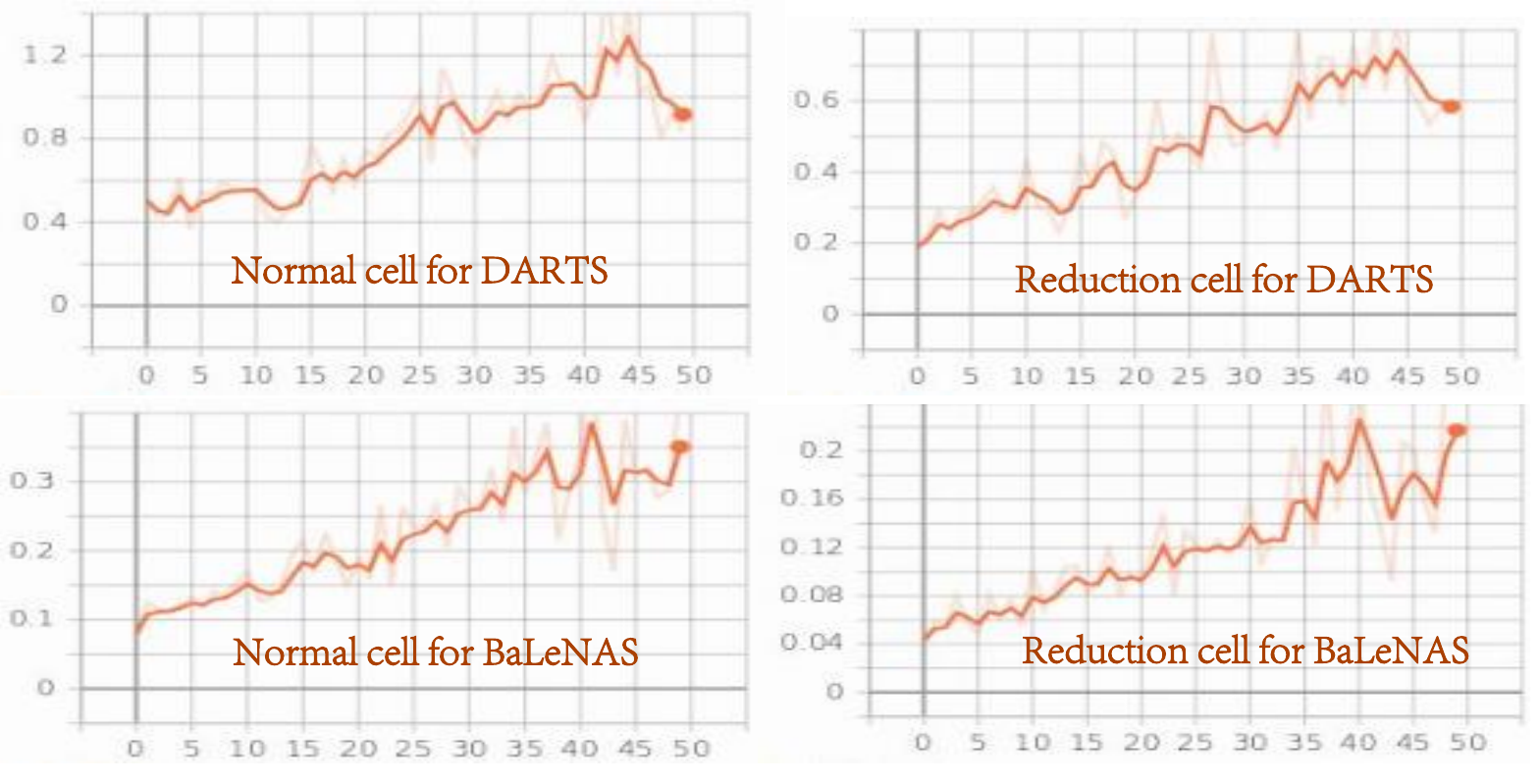}
  \caption{Trajectory of the Hessian norm in DARTS space.}
\label{fig:trace_norm}
\vspace{-2mm}
\end{figure}

\subsubsection{Tracking of the Hessian norm}
As described in Section \ref{sec2.1}, a large Hessian norm deteriorate the robustness of DARTS, and the incongruence between $\mathcal{L}_{val}(w^*,\alpha_{\theta}^*)$ and $\mathcal{L}_{val}(w^*,\alpha^*)$ is not negligible if we could not maintain the maintains the Hessian norm at a low level. The analysis in Sec. \ref{sec3.2} and Eq. \eqref{eq:stabilizing} shows that the loss function of the proposed BaLeNAS implicitly controls the trace norm of $\mathcal{H}$ similar as \cite{chen2020stabilizing,chen2020drnas}, helping stabilizing differentiable NAS. We plot the trajectory of the Hessian norm of BaLeNAS compared with the vanilla DARTS in Fig. \ref{fig:trace_norm}. As show, the Hessian norm in our BaLeNAS is always kept in a low level. Although the Hessian norm of BaLeNAS also increases with the supernet training similar as DARTS, BaLeNAS's largest Hessian norm is still smaller than DARTS in the early stage, showing the effectiveness of implicit regularization of our BaLeNAS as described in Sec. \ref{sec3.2}.

\section{Conclusion}
In this paper, we have formulated the architecture optimization in the differentiable NAS as a distribution learning problem and introduced a Bayesian learning rule to optimize the architecture parameters posterior distributions. We have theoretically demonstrated that the proposed framework can enhance the exploration for differentiable NAS and implicitly impose regularization on the Hessian norm to improve the stability. The above properties show that reformulating differentiable NAS as distribution learning is a promising direction. In addition, with leveraging the training-free proxies, our BaLeNAS can select more competitive architectures from the optimized distributions instead of applying \textit{argmax} on the mean to get the the discrete architecture, so that alleviate the discretization instability and enhance the performance. We operationalize the framework based on the common differentiable NAS baseline, DARTS, and experimental results on NAS benchmark datasets and the common DARTS search space have verified the proposed framework's effectiveness. 

Although BaLeNAS improves the differentiable NAS baseline by large margins, it computational consumption and memory consumption are similar with DARTS where our BaLeNAS is built on. Further questions include how to further decrease the computational and memory cost and also eliminate the \textit{depth gap} existing between architecture search and evaluation in differentiable NAS \cite{chen2019progressive}.


{\small
\bibliographystyle{ieee_fullname}
\bibliography{egbib}
}

\vspace{6cm}

\begin{table*}[ht]
\centering
\caption{Zero-cost NAS and FreeDARTS with different saliency metrics on NAS-Bench-201.}
\setlength{\tabcolsep}{3pt}
\begin{tabular}
{lcccccccc}
\toprule

{\multirow{2}*{Method}}&{Sample}&\multicolumn{2}{c}{CIFAR-10}&\multicolumn{2}{c}{CIFAR-100}&\multicolumn{2}{c}{ImageNet-16-120}\\
~&Size&\multicolumn{1}{c}{Valid(\%)}&\multicolumn{1}{c}{Test(\%)}&\multicolumn{1}{c}{Valid(\%)}&\multicolumn{1}{c}{Test(\%) }&\multicolumn{1}{c}{Valid(\%) }&\multicolumn{1}{c}{Test(\%)}\\
\midrule
BaLeNAS&- &91.32$\pm$0.09&94.02$\pm$0.14&71.53$\pm$0.08&71.93$\pm$0.27&45.39$\pm$0.17&45.48$\pm$0.39\\
\multirow{3}*{BaLeNAS with SNIP}&10&90.95$\pm$0.39&93.85$\pm$0.22&71.37$\pm$0.35&71.48$\pm$0.52&46.04$\pm$0.47&46.03$\pm$0.41\\
~&50&88.23$\pm$2.18&92.56$\pm$1.18&68.26$\pm$2.74&64.58$\pm$3.18&27.13$\pm$9.20&35.23$\pm$10.3\\
~&100&86.04$\pm$0.00&91.37$\pm$0.00&65.52$\pm$0.00&67.77$\pm$0.00&36.33$\pm$0.00&24.97$\pm$0.00\\
\multirow{3}*{BaLeNAS with Grasp}&10&91.10$\pm$0.23&93.94$\pm$0.05&72.03$\pm$0.53&72.00$\pm$0.06&45.26$\pm$0.56&44.67$\pm$1.54\\
~&50&90.56$\pm$0.76&93.72$\pm$0.16&71.52$\pm$1.03&70.62$\pm$1.43&45.01$\pm$0.81&44.92$\pm$1.64\\
~&100&89.01$\pm$0.78&92.32$\pm$1.25&67.86$\pm$2.61&67.32$\pm$1.85&40.29$\pm$3.91&39.84$\pm$3.43\\
\multirow{3}*{BaLeNAS with SynFlow}&10&91.52$\pm$0.04&94.08$\pm$0.13&72.37$\pm$0.53&72.55$\pm$0.42&45.34$\pm$0.23&45.82$\pm$0.30\\
~&50&91.52$\pm$0.04&94.33$\pm$0.03&72.67$\pm$0.41&72.95$\pm$0.28&46.14$\pm$0.23&46.54$\pm$0.36\\
~&100&91.52$\pm$0.04&94.33$\pm$0.03&72.67$\pm$0.41&72.95$\pm$0.28&46.14$\pm$0.23&46.54$\pm$0.36\\
\bottomrule
\end{tabular}
\label{tab:different_proxies}
\end{table*}

\section*{APPENDIX:}

\subsection*{A. Search Spaces and Experimental Setting}

In our experiments, we consider two scenarios, NAS benchmark datasets and the common DARTS space, to analyze the proposed framework FreeDARTS. The high computational cost in evaluation is a major obstacle when analyzing and reproducing differentiable NAS methods. To alleviate this issue, several benchmark datasets have been recently published \cite{ying2019bench,BENCH102,zela2020nasbench1shot1,siems2020bench}, where the ground-truth for all candidate architectures in the benchmark datasets is known. The NAS-Bench-201 dataset \cite{BENCH102} is a popular NAS benchmark dataset to analyze differentiable NAS methods. The search space in NAS-Bench-201 contains four nodes with five associated operations, resulting in 15,625 cell candidates, where the performance of CIFAR-100, CIFAR-100, and ImageNet for all architectures in this search space are reported. The NAS-Bench-101 \cite{ying2019bench} is another famous NAS benchmark dataset, which is much larger than NAS-Bench-201 while only the CIFAR-10 performance for all architectures are reported. More important, the architectures in NAS-Bench-101 contain different number of nodes, which makes it impossible to build a generalized supernet for one-shot nor differential NAS methods. To leverage the NAS-Bench-101 for analyzing the differentiable NAS methods, NAS-Bench-1shot1 \cite{zela2020nasbench1shot1} builds from the NAS-Bench-101 benchmark dataset by dividing all architectures in NAS-Bench-101 into 3 different unified cell-based search spaces, which contain 6240, 29160, and 363648 architectures, respectively. The architectures in each search space have the same number of nodes and connections, making the differentiable NAS could be directly applied to each search space. We choose the third search space in NAS-Bench-1shot1 since it is much more complicated than the remaining two search spaces. 

\begin{figure}[t]
\centering
  \begin{minipage}{8cm}
      \includegraphics[width=8cm,height=2.5cm]{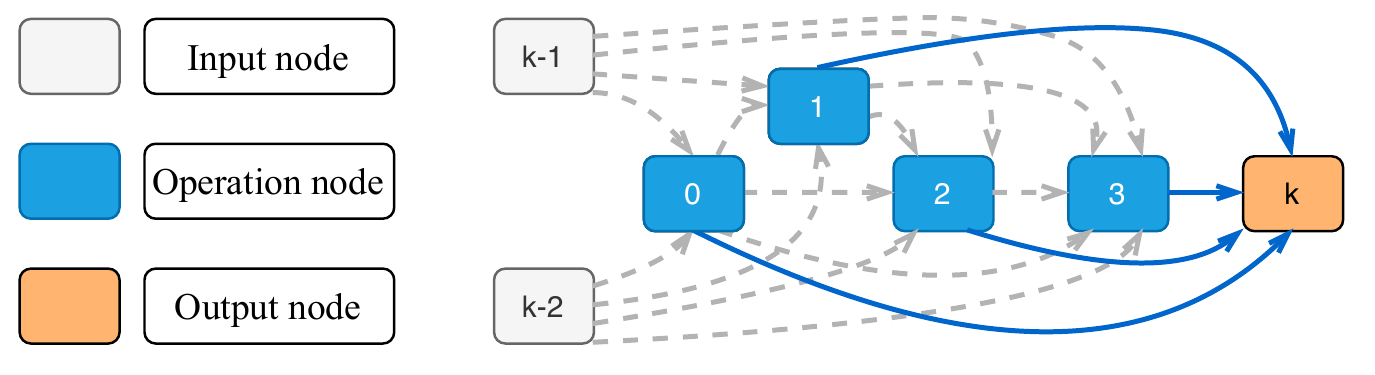}
  \end{minipage}
 \caption{Description of DARTS search space.}
 \label{fig:darts_search_space}
\end{figure}

\begin{table*}[ht]
\centering
\caption{Ablation study on the MCMC sampling size on NAS-Bench-201. 
}
\begin{tabular}
{|l|c|c|c|c|c|c|c|c|}
\hline

{\multirow{2}*{MCMC number}}&\multicolumn{2}{c|}{CIFAR-10}&\multicolumn{2}{c|}{CIFAR-100}&\multicolumn{2}{c|}{ImageNet-16-120}\\
~&\multicolumn{1}{c}{Valid(\%)}&\multicolumn{1}{c|}{Test(\%)}&\multicolumn{1}{c}{Valid(\%)}&\multicolumn{1}{c|}{Test(\%) }&\multicolumn{1}{c}{Valid(\%) }&\multicolumn{1}{c|}{Test(\%)}\\
\hline
\hline
$M=1$&90.52$\pm$0.09&93.33$\pm$0.04&70.67$\pm$0.08&70.95$\pm$0.27&44.39$\pm$0.47&44.32$\pm$0.39\\
$M=2$&90.71$\pm$0.12&93.75$\pm$0.87&71.25$\pm$0.92&71.43$\pm$0.45&44.63$\pm$0.55&45.05$\pm$0.95\\
$M=3$&91.32$\pm$0.09&94.02$\pm$0.14&71.53$\pm$0.08&71.93$\pm$0.27&45.39$\pm$0.17&45.48$\pm$0.39\\
$M=4$&90.03$\pm$0.96&93.04$\pm$1.09&68.80$\pm$1.46&69.20$\pm$1.86&43.09$\pm$2.93&43.21$\pm$2.88\\
\hline
Random baseline&83.20$\pm$13.28&86.61$\pm$13.46&60.70$\pm$12.55&60.83$\pm$12.58&33.34$\pm$9.39&33.13$\pm$9.66\\
DARTS (2nd)&37.51$\pm$3.19&53.89$\pm$0.58&13.37$\pm$2.35&13.96$\pm$2.33&15.06$\pm$1.95&14.84$\pm$2.10\\
\textbf{optimal}&91.61&94.37&74.49&73.51&46.77&47.31\\
\hline
\end{tabular}
\flushleft{}
\label{tab:M_ablation}
\end{table*}

As to the most common search space in NAS, DARTS needs to search for two types of cells: a normal cell $\alpha_{normal}$ and a reduction cell $\alpha_{reduce}$. Cell structures are repeatedly stacked to form the final CNN structure. There are only two reduction cells in the final CNN structure, located in the 1/3 and 2/3 depths of the network. There are seven nodes in each cell: two input nodes, four operation nodes, and one output node. Each operation node selects two of the previous nodes' output as input nodes in this search space. Each input node will select one operation from $|\mathcal{O}|=8$ candidate operations. Fig. \ref{fig:darts_search_space} describes a unified convolutional search space in DARTS. The common practice in DARTS is to search on CIFAR-10, and the best searched cell structures are directly transferred to CIFAR-100 and ImageNet. The experimental settings on DARTS space in this paper are following the common DARTS setting. We conduct the architecture search with 5 different \textit{random seeds}, and the best one is selected after the evaluation on CIFAR-10, which is then transferred to CIFAR-100 and ImageNet. The architecture evaluation for CIFAR-10 and CIFAR-100 are on a single GPU with batch size 96, while for ImageNet is performed on 2 GPUs. Since the sizes of searched architectures are in a range, we adjust the number of filter in the evaluation to make the model sizes similar for fair comparison. We use a linear learning rate scheduler with following PDART \cite{chen2019progressive} and PCDARTS \cite{xu2019pcdarts} to use a smaller slope in the last five epochs for the architecture evaluation on ImageNet \footnote{The reproducible codes could be found in the supplementary material. }.

\subsection*{B. Ablation Study on the Saliency Metrics} 
In our BaLeNAS-TF, we utilize three train-free saliency metrics, SNIP, GraSP, and Synflow, as proxies for the architecture selection from the optimized distribution. In Table \ref{tab:different_proxies}, we considered different number of sample size for our BaLeNAS-TF when combined with the three saliency metrics. As shown in Table \ref{tab:different_proxies}, combined with different train-free proxies, our BaLeNAS-TF achieve higher performance than the original BaLeNAS when the sample size is 10. However, when increasing the sample size, we can see a sharp drop for BaLeNAS-TF with SNIP and GraSP, showing the two metrics are not appropriate metrics to for the architecture selection. On the contrary, the SynFlow, also adopted by our BaLeNAS-TF, shows a clear improvement with the sample size from 10 to 100, implying that this proxies is more reliable for the architecture selection.

\subsection*{C. Ablation Study of MCMC on NAS-Bench-201}
As we described in Section \ref{sec3.2}, one key additional hyperparameter in BaLeNAS is the sampling number $M$ in MCMC, and this subsection investigates how this hyperparameter affects the performance of BaLeNAS. Table \ref{tab:M_ablation} summarizes the performance our BaLeNAS (2nd) with different number of MCMC sampling. As shown, our BaLeNAS is very robust to the  number of MCMC sampling, where BaLeNAS achieves excellent results under different scenarios, outperforming most existing NAS baselines. An interesting observation is that the performance of BaLeNAS increase with multiple samplings when $M<4$ in MCMC, and $M=3$ achieves the best performance. Theorem 1 in \cite{khan2018fast} points out that VAdam with $M>1$ will converge fast while might result in slightly less exploration. The exploration and exploitation can be balanced by the MCMC sample size. A detailed explanation can be found in the Section 3.4 of \cite{khan2018fast}.


\begin{figure}
 \subfloat[Normal cell of BaLeNAS]{
  \begin{minipage}{4.2cm}
      \includegraphics[width=4.2cm,height=3cm]{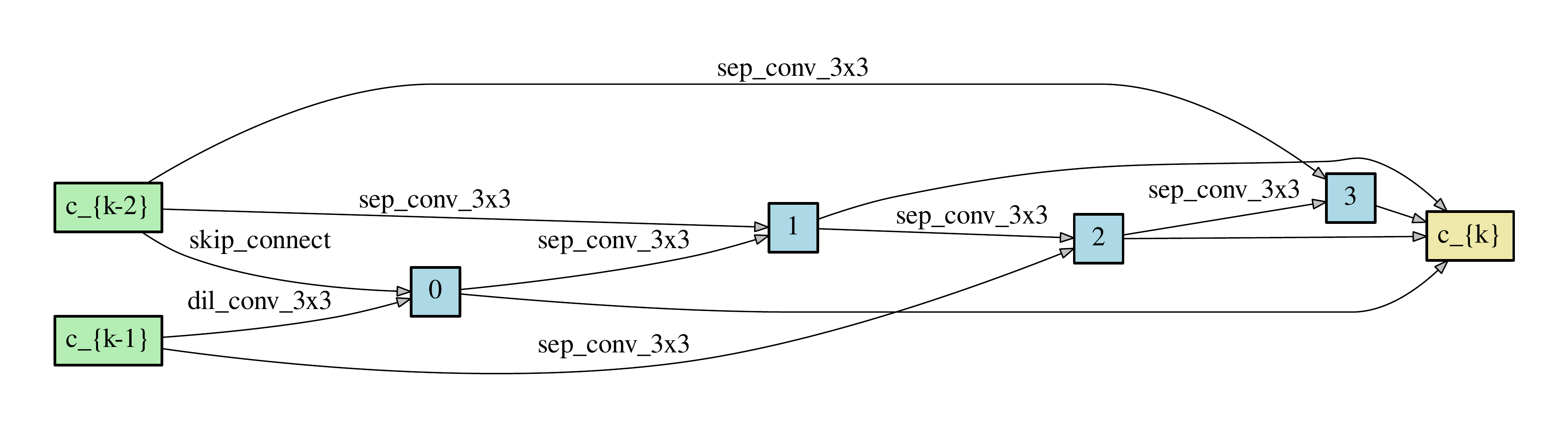}
  \end{minipage}
 }
  \subfloat[Reduction cell of BaLeNAS]{
  \begin{minipage}{4.2cm}
       \includegraphics[width=4.2cm,height=3cm]{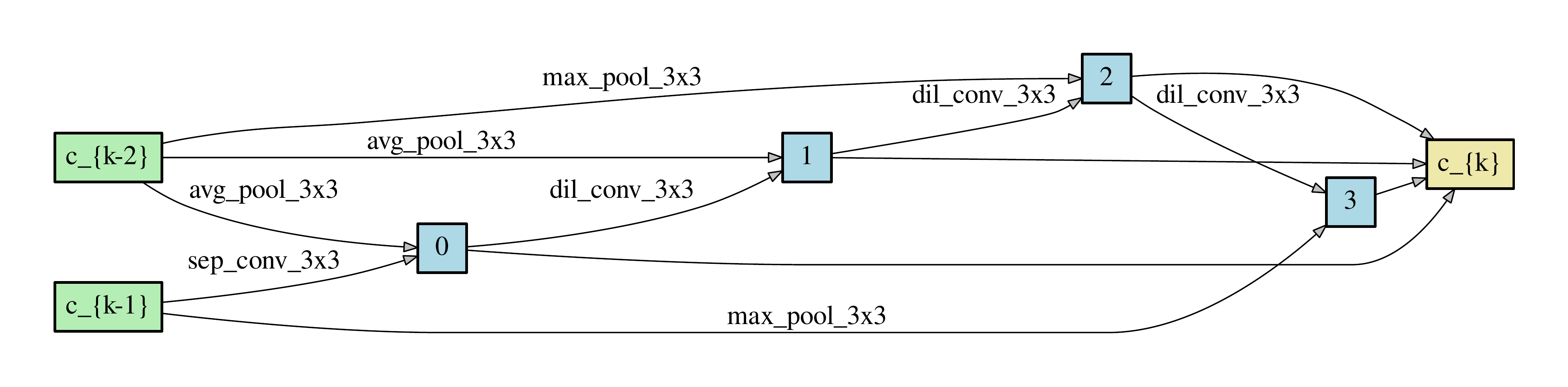}
  \end{minipage} 
  }
  
 \subfloat[Normal cell of BaLeNAS w/o]{
  \begin{minipage}{4.2cm}
      \includegraphics[width=4.2cm,height=3cm]{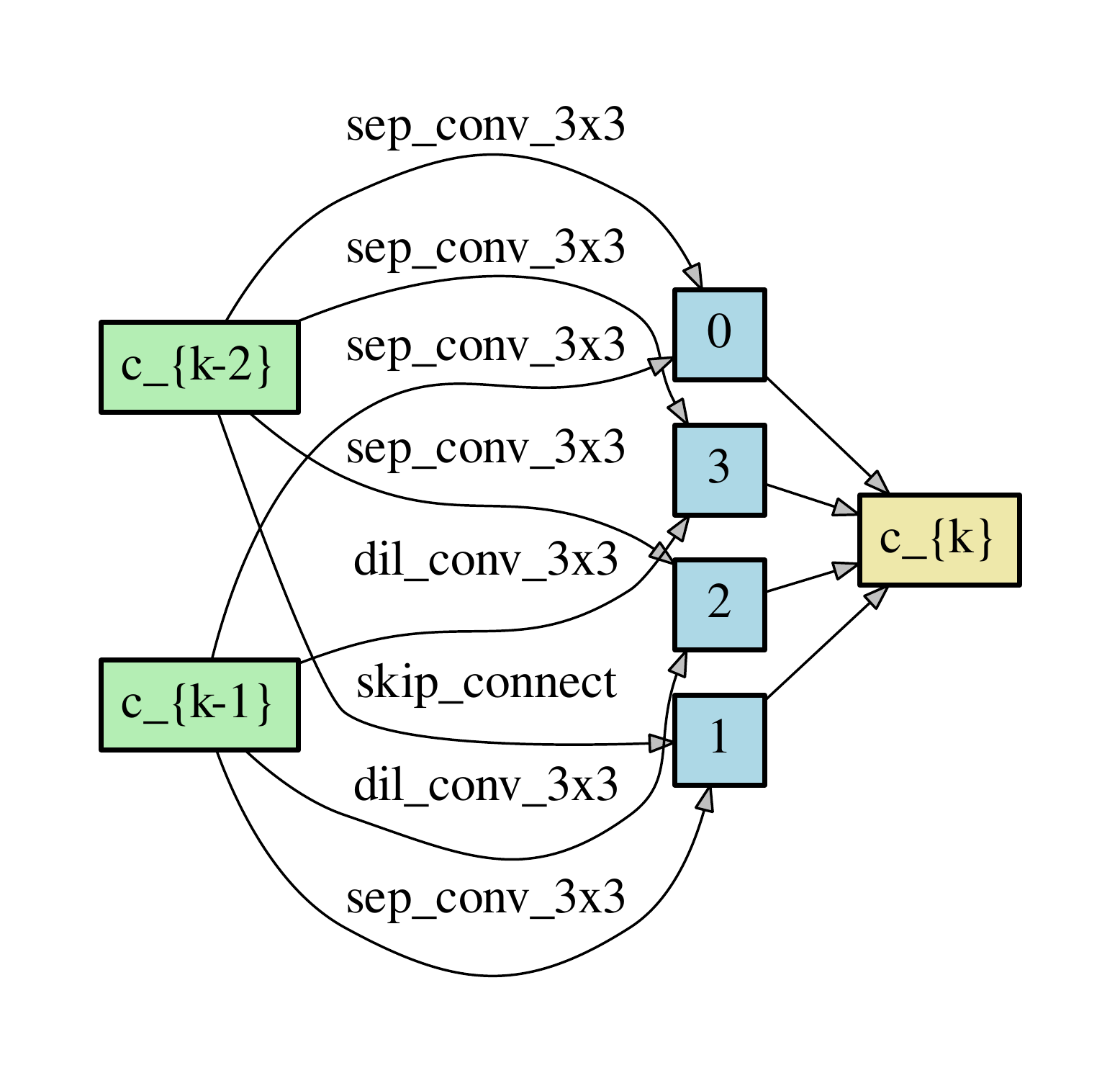}
  \end{minipage}
 }
  \subfloat[Reduction cell of BaLeNAS w/o]{
  \begin{minipage}{4.2cm}
       \includegraphics[width=4.2cm,height=3cm]{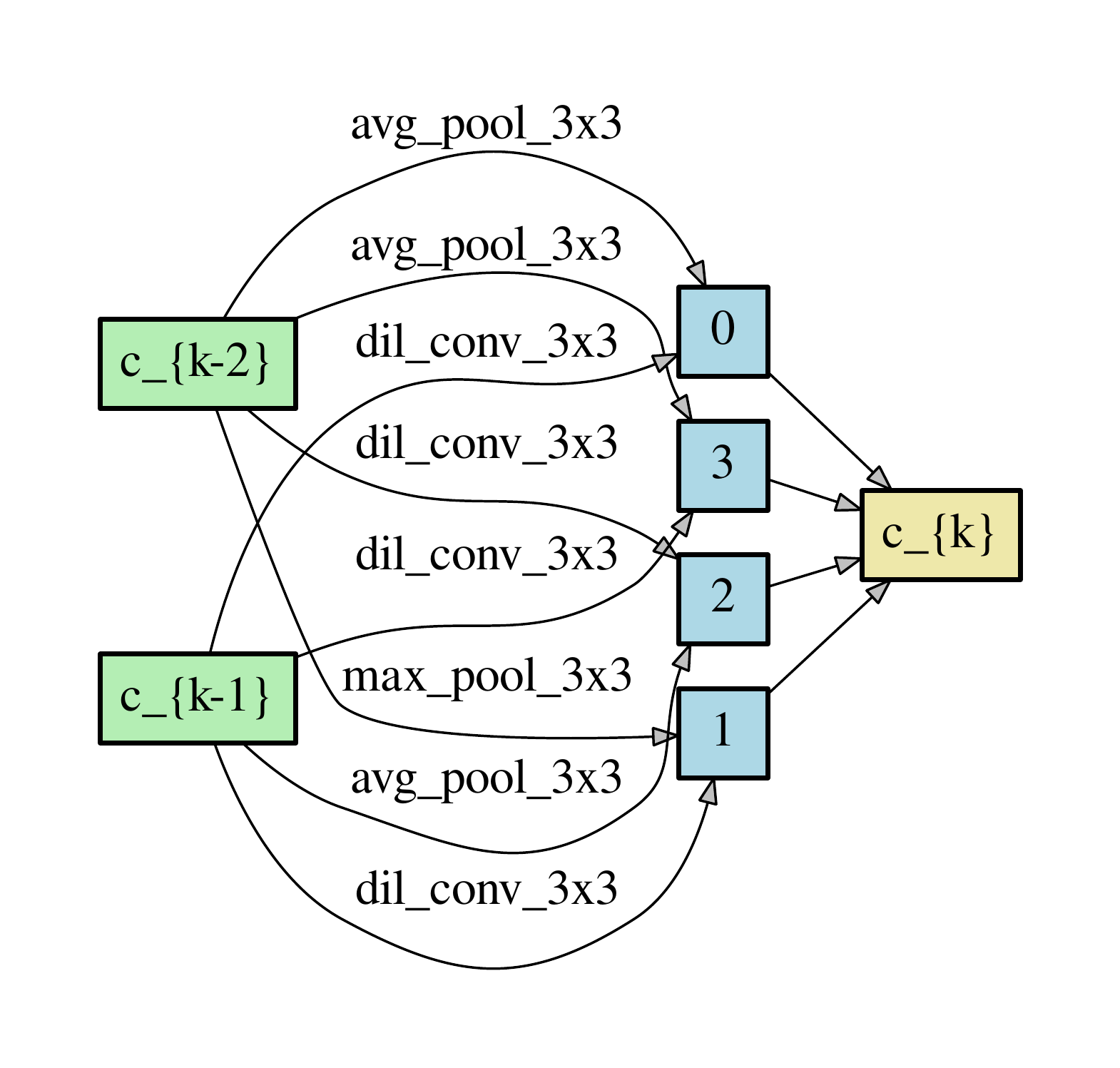}
  \end{minipage} 
  }

 \caption{Examples of searched cells by BaLeNAS and BaLeNAS without regularizatons (BaLeNAS w/o).}
 \label{fig:cellexamples}
\end{figure}

\subsection*{D. Searched Architectures Visualization}
Fig.\ref{fig:cellexamples} plots the searched architectures on DARTS space by BaLeNAS and BaLeNAS-TF. We could observe that, our BaLeNAS tends to obtain ``shallow" architectures, which is also observed by several existing works \cite{zhang2020onetpami,nayman2019xnas}. As we know the shallow architectures are easier to train and usually perform excellently in the small dataset, implying that the differentiable NAS methods prefer those ``shallow" architectures if we only utilize the validation accuracy as the indicator. However, the performance of those ``shallow" architectures on the large dataset is not as competitive as on the small dataset, indicating poor transferability. These results suggest the importance of introducing other indicator to differentiable NAS for architecture search, especially in the complicated real-world search space, to help finding more robust architectures. In contrast, as shown in Fig.\ref{fig:cellexamples}, our BaLeNAS-TF can found ``deeper" architectures as it does not only rely on the validation accuracy for the architecture, but also another saliency metric. We can find a similar phenomenon in the NAS-Bench-201 search space that, even though DrNAS achieves near-optimal results on CIFAR-10, while our BaLeNAS-TF outperform it on the larger dataset.

\end{document}